\documentclass[10pt,twocolumn,letterpaper]{article}

\usepackage[pagenumbers]{iccv} 
\usepackage{multirow}
\usepackage{booktabs}
\usepackage{enumitem}
\usepackage{ulem}
\usepackage{microtype}
\usepackage{wasysym}
\usepackage{colortbl}
\usepackage[numbers,sort&compress]{natbib}
\definecolor{iccvblue}{rgb}{0.21,0.49,0.74}
\usepackage[pagebackref,breaklinks,colorlinks,allcolors = iccvblue]{hyperref}

\def\paperID{3924} 
\def\confName{ICCV}
\def\confYear{2025}

\title{SpikeDerain: Unveiling Clear Videos from Rainy Sequences \\ Using Color Spike Streams}

\author{Hanwen~Liang$^{1,\#}$ \quad Xian~Zhong$^{1,}$\thanks{\ Corresponding author. \# Contributed equally to this work.} \quad Wenxuan~Liu$^{2,1,\#}$ \quad Yajing~Zheng$^2$ \quad Wenxin~Huang$^3$ \\ Zhaofei~Yu$^2$ \quad Tiejun~Huang$^2$ \\
$^1$ Wuhan University of Technology \quad 
$^2$ Peking University \quad
$^3$ Hubei University \\
{\tt\small \{lianghanwen, zhongx\}@whut.edu.cn, \{liuwx666, yj.zheng\}@pku.edu.cn,}, \\ 
{\tt\small wenxinhuang\_wh@163.com, \{yuzf12, tjhuang\}@pku.edu.cn}
}

\begin{document}
\maketitle

\begin{abstract}
Restoring clear frames from rainy videos presents a significant challenge due to the rapid motion of rain streaks. Traditional frame-based visual sensors, which capture scene content synchronously, struggle to capture the fast-moving details of rain accurately. In recent years, neuromorphic sensors have introduced a new paradigm for dynamic scene perception, offering microsecond temporal resolution and high dynamic range. However, existing multimodal methods that fuse event streams with RGB images face difficulties in handling the complex spatiotemporal interference of raindrops in real scenes, primarily due to hardware synchronization errors and computational redundancy. In this paper, we propose a Color Spike Stream Deraining Network (SpikeDerain), capable of reconstructing spike streams of dynamic scenes and accurately removing rain streaks. To address the challenges of data scarcity in real continuous rainfall scenes, we design a physically interpretable rain streak synthesis model that generates parameterized continuous rain patterns based on arbitrary background images. Experimental results demonstrate that the network, trained with this synthetic data, remains highly robust even under extreme rainfall conditions. These findings highlight the effectiveness and robustness of our method across varying rainfall levels and datasets, setting new standards for video deraining tasks.
The code will be released soon.

\end{abstract}

\begin{figure}
	\centering
	\includegraphics[width = \linewidth]{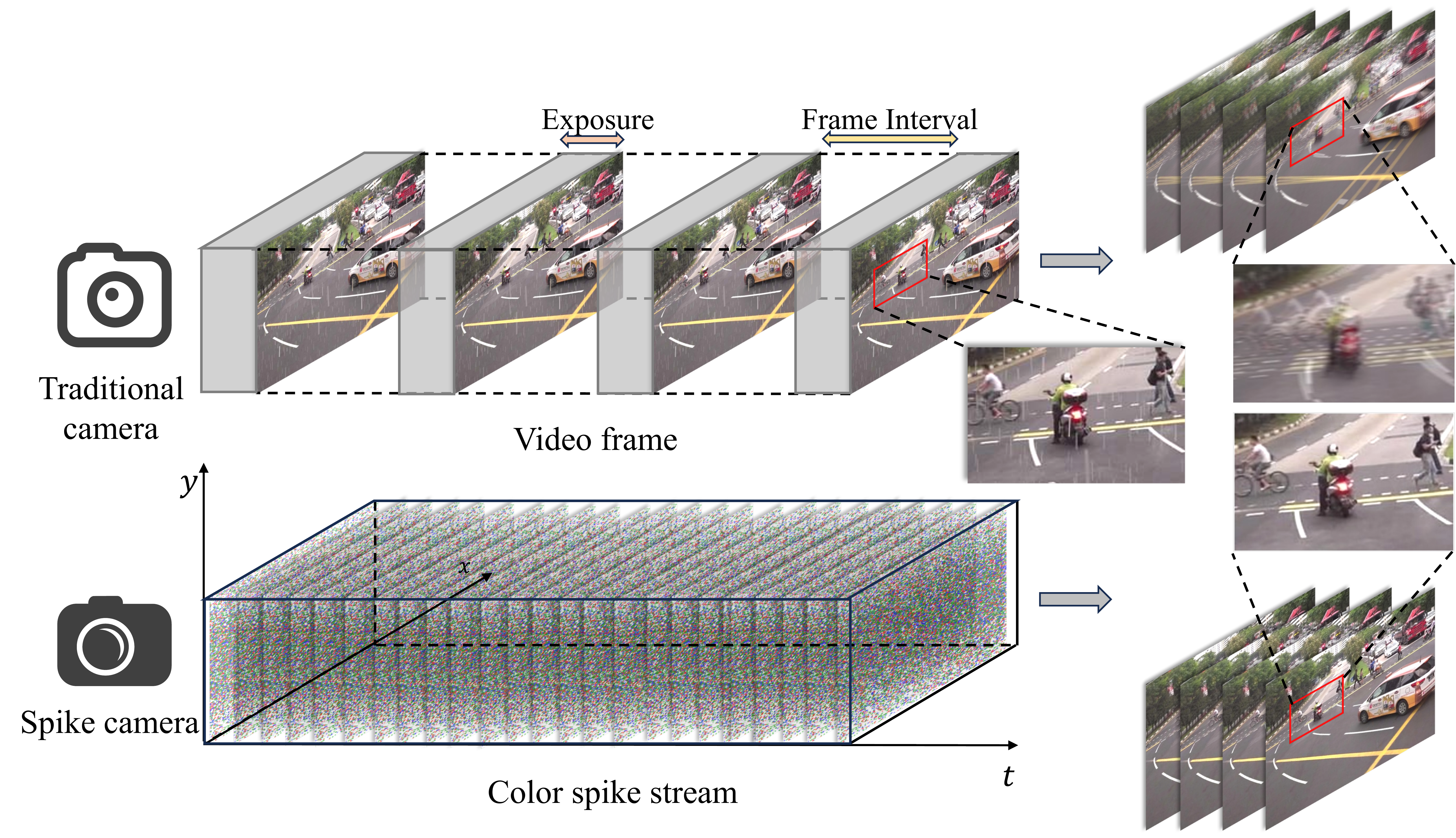}
	\vspace{-15pt}
	\caption{\textbf{Comparison of Information Capture between Traditional Video Frames and Spike Streams.} Traditional cameras, due to fixed exposure time and sampling frequency, introduce dynamic blur and inter-frame intervals, which degrade the image quality for subsequent tasks. In contrast, spike streams, with their ultra-high temporal resolution, capture high-speed information more effectively.}
	\label{fig:motivation}
\end{figure}

\section{Introduction}
\label{sec:intro}

Rainfall is one of the most common severe weather phenomena, leading to significant degradation in captured videos and images. This degradation not only reduces visual quality, but also significantly impacts the reliability of computer vision tasks that rely on clear input, such as object detection~\cite{40} and person re-identification~\cite{tcsv/ZhongLHYJL22}. As shown in \cref{fig:motivation}, traditional RGB cameras are limited by fixed exposure mechanisms~\cite{42}, making it difficult to capture the transient motion of high-speed raindrops. This limitation presents a fundamental bottleneck in recovering details in existing frame-based video deraining methods. In recent years, neuromorphic sensors, such as event cameras~\cite{43,45,event-Pose-Estimation}, have provided a new paradigm for dynamic scene perception, offering microsecond temporal resolution and high dynamic range. However, two major challenges remain when reconstructing rain-free images from the high-frequency, asynchronous data stream of neuromorphic sensors: First, the event stream only records sparse events of brightness changes~\cite{26}, losing absolute brightness and color information, making it difficult to distinguish between rain streaks and background textures. Second, existing multimodal methods~\cite{46} that fuse event streams with RGB images struggle to handle the complex spatiotemporal interference of raindrops in real scenes due to hardware synchronization errors~\cite{47} and computational redundancy.



To address these challenges, we propose the first spike-driven single-modal rain removal framework. Unlike event cameras, spike cameras generate spike streams by continuously capturing the absolute brightness of each pixel~\cite{14,17,20,24}, and spike cameras with a Color Filter Array (CFA)~\cite{23,48,49,50} retain certain color information while acquiring spike streams. This natural fusion of spatiotemporal continuity and multispectral information offers dual advantages for rain removal tasks: \textit{Physical consistency}, the single-sensor color spike stream eliminates cross-modal alignment errors in high-speed scenes; and \textit{Information integrity}, where the spike sequence preserves both dynamic details and accurate color distribution, mitigating rain streak-background confusion seen in traditional methods.

In this paper, we design a spike-driven deraining network ({SpikeDerain}) that maps raw spike streams into a continuous spatiotemporal representation through spike density and color correlation, optimizing both image reconstruction and rain streak separation processes. To complement our single-modal spike-driven framework, we introduce a synthetic dataset that addresses two critical challenges: the lack of dynamic continuous rain streaks and the loss of background consistency inherent in event camera data. Our physically interpretable rain streak synthesis model generates parameterized, continuously evolving rain streaks that are well aligned with the high temporal resolution of spike cameras. Experiments show that the network, trained with this synthetic data, remains highly robust under extreme rainfall conditions. 

The contributions can be summarized fourfold:

\begin{itemize}
	\item We utilize the spatiotemporal joint characterization capability of color spike streams to achieve spike-driven single-modal rain removal for the first time.

	\item We propose a joint optimization deraining network driven by color spike streams, capable of reconstructing dynamic scenes and accurately removing rain streaks.

	\item We design a parameterized rain streak generator that quantitatively controls the physical properties of rain streaks.

	\item We achieve significant improvements compared to state-of-the-art methods, with extensive experiments demonstrating the effectiveness of each module.

\end{itemize}

\section{Motivation}

Traditional rain removal methods often rely on frame-based processing~\cite{9,S2VD,ESTIL}, while spike stream and event stream, as emerging forms of dynamic perception data, provide new ideas~\cite{43,59,Optical-Flow-Guided-Object-Pose-Tracking}. Event cameras and spike cameras use different technological principles when capturing dynamic scenes, and there are significant differences between the two in terms of information capture methods and performance capabilities.

\begin{figure}
	\centering
	\includegraphics[width = \linewidth]{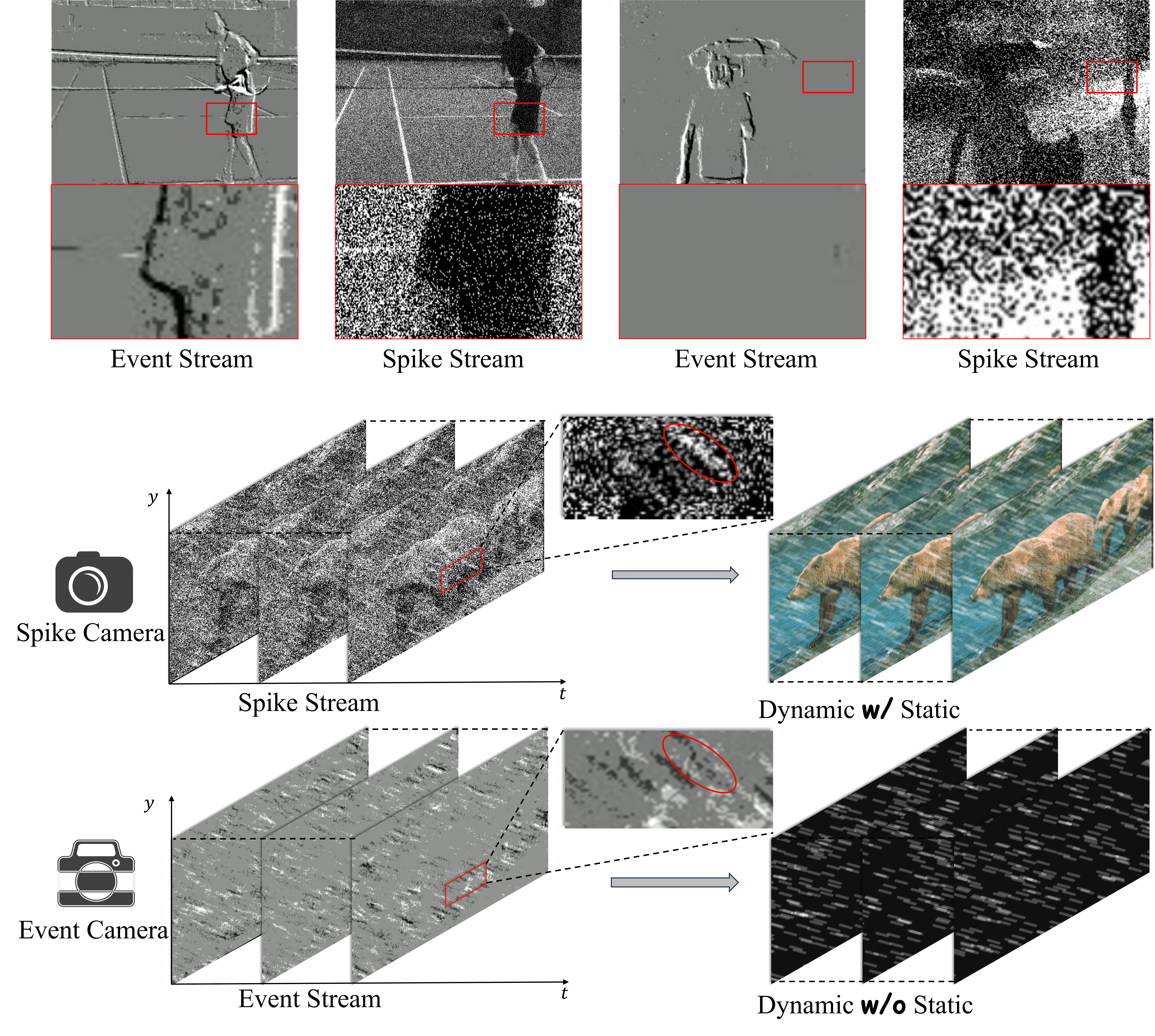}
	\vspace{-15pt}
	\caption{\textbf{Difference between Event-based and Our Spike-based Rainy Removal.} Event cameras only record relative light intensity. In contrast, spike cameras record absolute light intensity, thereby preserving essential static information.}
	\label{fig:event_spike}
\end{figure}

Unlike traditional cameras, event cameras have ultra-high temporal resolution and can capture high-speed motion information by responding to relative changes in pixel brightness. However, the event camera only records sparse events of brightness changes~\cite{45,event-pose-tracking}, and its output lacks absolute brightness and color information of the scene. As shown in \cref{fig:event_spike}, due to the loss of absolute brightness, completely different objects in the event stream data may visually exhibit the same properties, thereby losing more background details. In addition, due to the unique imaging mechanism of event cameras, only changing pixels are recorded in the time stream, resulting in a significant loss of static information when dealing with fixed-angle scenes. To address these issues, existing research typically employs multimodal inputs~\cite{46}. However, this leads to complex problems such as data alignment, hardware synchronization~\cite{47}, and computational redundancy.

In contrast, spike cameras continuously record the absolute brightness information of each pixel~\cite{17,24,30}, generating a spike stream that not only preserves the dynamic details brought by high temporal resolution but also fully preserves the light intensity and color distribution of the scene~\cite{23}. This feature makes the spike camera have obvious advantages in video deraining tasks: First, it delivers comprehensive scene information in a single modality, eliminating the need for supplementary sensors; second, its rich absolute brightness data enables precise raindrop-background separation, ensuring superior background reconstruction.

\begin{figure*}
	\centering
	\includegraphics[width = \linewidth]{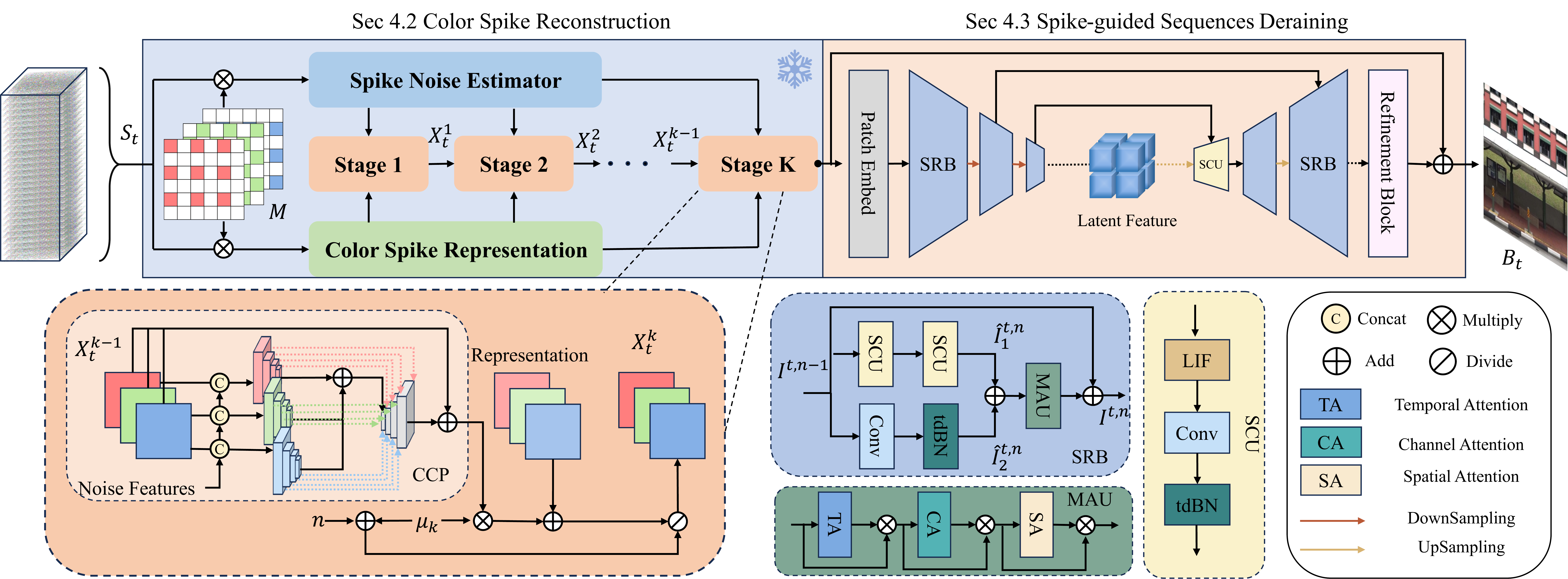}
	\vspace{-15pt}
	\caption{\textbf{Architecture of Proposed Spike-Driven Deraining Network (SpikeDerain).} Our method leverages color spike streams from a spike camera by integrating a color spike reconstruction module with a multi-stage SNN-based deraining module to efficiently restore rain-free videos.}
	\label{fig:framework}
\end{figure*}

\section{Related Work}
\label{sec:formatting}


\subsection{Video Deraining}

Video deraining differs from single-image deraining by exploiting the temporal continuity inherent in video sequences. Early work by Garg and Nayar~\cite{1,2,3,4} introduced a rain streak removal method based on linear spatiotemporal modeling. Subsequently, researchers improved video deraining performance using various priors, including Fourier domain techniques~\cite{5}, geometric structures~\cite{6}, directional features~\cite{7}, and patch-based methods~\cite{8}. Recent advancements in video deraining have been driven by deep learning technologies. 
For unsupervised training, Yang \textit{et al.}~\cite{11} introduced a self-learning method, and Yue \textit{et al.}~\cite{12} further improved video deraining by employing semi-supervised learning with a dynamic generator. 
In this work, we bring in a spike camera, avoiding the challenges of alignment and the computational burden associated with cross-modality.

\subsection{Spike-Based Vision}

Spike cameras~\cite{14,15,16,17} are inspired by the primate retina and differ from traditional cameras by generating synchronized spike streams for each pixel with extremely low latency. Unlike event cameras, spike cameras record absolute light intensity, which enables better recovery of texture details. With their ultra-high temporal resolution and low energy consumption, spike cameras are widely applied in various tasks, including object detection~\cite{21,22}, super-resolution~\cite{23,24,25}, dynamic deblurring~\cite{26,27}, and motion estimation~\cite{28}. These works introduce a spike camera as an innovative sensor that provides complementary information, achieving significant progress. In this paper, we make the first attempt to investigate the role of spike cameras in video deraining.


\subsection{Spiking Neural Networks}

SNNs mimic biological neurons by processing information through discrete binary spikes over time, offering energy-efficient computation suited for neuromorphic hardware~\cite{32,33}. While many SNNs~\cite{Youhong,Hu_shengwang} are created by converting pre-trained artificial neural networks (ANNs) their performance remains constrained by the original ANN's architecture. Recent advances in directly training SNNs bypass this limitation~\cite{37}, achieving competitive accuracy with fewer computational steps. Techniques like surrogate gradients~\cite{36} and biologically inspired learning rules enable efficient backpropagation, while architectural innovations (\textit{e.g.}, residual connections adapted for spikes~\cite{39}) address vanishing gradients and deepen SNNs effectively. In this work, we propose a directly training SNNs to address video deraining.




\section{Proposed Method}

We first introduce the working mechanism of the spike camera and the mathematical model of the video deraining task, followed by the reconstruction of the color spike stream and the video deraining module. The overall framework of our method is shown in \cref{fig:framework}.

\begin{figure}
	\centering
	\includegraphics[width = 0.9\linewidth]{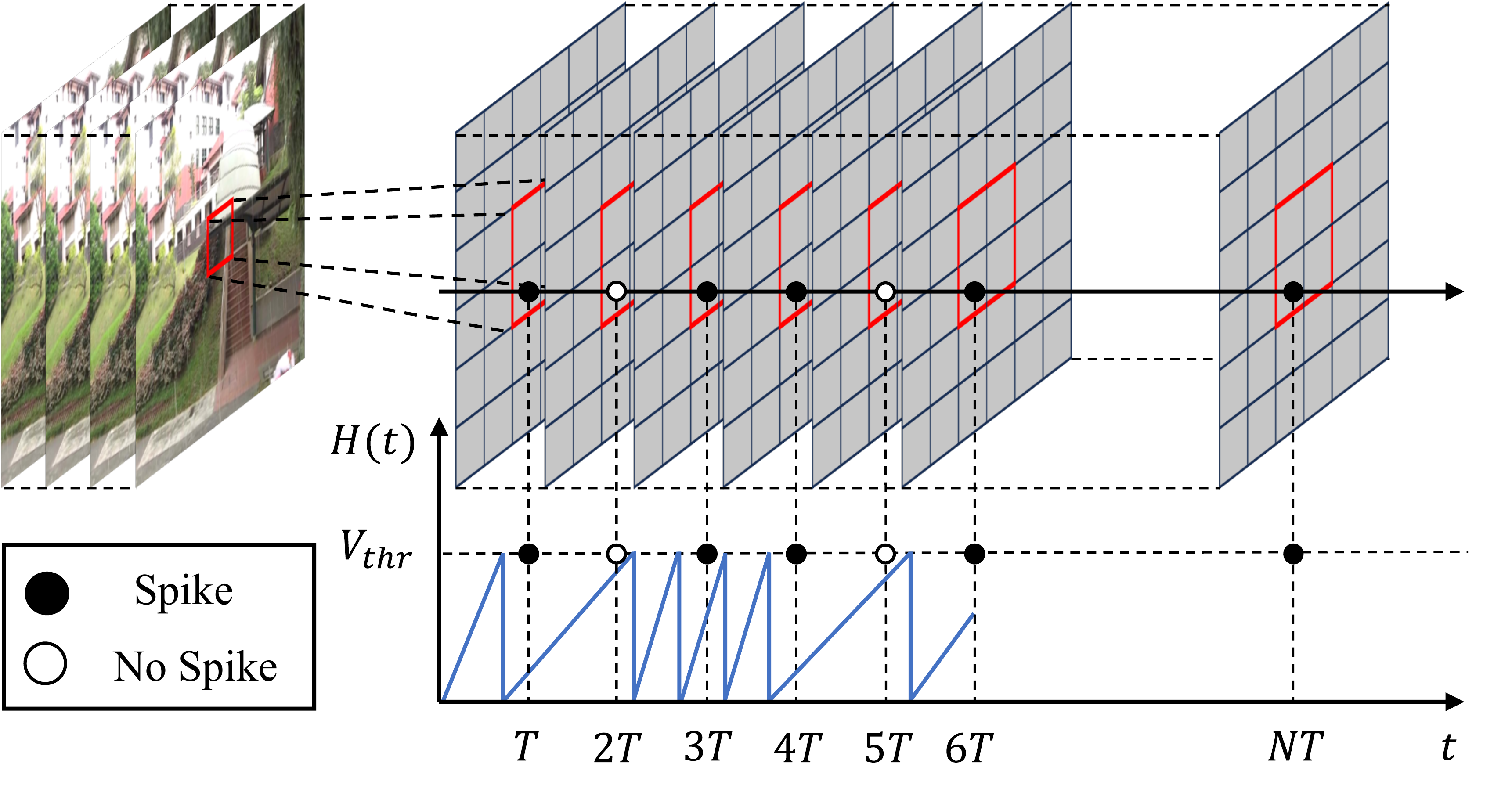}
	\vspace{-5pt}
	\caption{\textbf{Spike Camera Mechanism.} Each pixel continuously monitors light intensity and emits a binary spike once a preset brightness threshold is reached, enabling ultra-high temporal resolution capture of dynamic scenes.}
	\label{fig:spike_camera}
\end{figure}

\subsection{Preliminaries}

\paragraph{Spike Camera Mechanism.} The Spike Camera~\cite{17} is a neuromorphic sensor that differs fundamentally from traditional RGB and event cameras. In \cref{fig:spike_camera}, each photon detector continuously measures the brightness and emits a binary spike once the integrated photon count exceeds a preset threshold, after which the integration circuit resets. This generates an asynchronous spike stream $S(x,y,t)\in\{0,1\}^{H \times W \times 1 \times T}$, where each spike encodes the position $(x,y)$, timestamp $t$, and the absolute brightness response of the scene $I$. The mathematical formulation is as follows:
\begin{equation}
	S(x,y,t) = 
	\begin{cases}
	1 & \int_{t_0}^t I \left( x,y,\tau \right) d\tau \geq \theta, \\
	0 & \mathrm{otherwise}, 
	\end{cases}
\end{equation}
where $I(x,y,\tau)$ is the instantaneous light intensity, and $\theta$ is the spike trigger threshold.



\vspace{-10pt}
\paragraph{Problem Formulation.} Conventional video deraining methods process continuous frames $I = \{I_t\}_{t = 1}^n$, where each frame is modeled as the sum of a clean background $B_t$ and a rain streak layer $R_t$:
\begin{equation}
	I_t = B_t + R_t.
\end{equation}
In contrast, a spike camera produces a spatiotemporal binary spike stream $S \in \{0,1\}^{H \times W \times 1 \times T}$, where $H$ and $W$ denote the image height and width, and $T$ represents the length of the spike sequence. Spike stream generation can be modeled as follows:
\begin{equation}
	S = \mathcal{G} \left( B + R \right) + N, \label{con:spike_raw}
\end{equation}
where $\mathcal{G}(\cdot)$ is the quantization function of the spike camera, mapping continuous brightness into a binary spike sequence, and $N$ represents the quantization noise, modeled as Poisson-Gaussian noise:
\begin{equation}
	N \sim \mathcal{N} \left( 0, \sigma_g^2 + \sigma_p^2 \cdot I_t \right).
\end{equation}
To mitigate both rain streak and quantization noise, we propose an optimization objective for joint spike reconstruction and rain removal:


\begin{equation}
	\min_{\Phi} \mathcal{L}_{\mathrm{derain}} \left( G_{\Phi} \left( F_{\Theta} \left( S \right) \right), B \right),
\end{equation}
where $F_{\Theta}(\cdot)$ represents the spike reconstruction module, which maps $S$ to an initial estimation; $G_{\Phi}$ is the deraining module, which separates the background from the rain layer. $\mathcal{L}_{\mathrm{recon}}$ and $\mathcal{L}_{\mathrm{derain}}$ represent the reconstruction loss and rain removal loss, respectively.

\begin{figure}
	\centering
	\includegraphics[width = 0.9\linewidth]{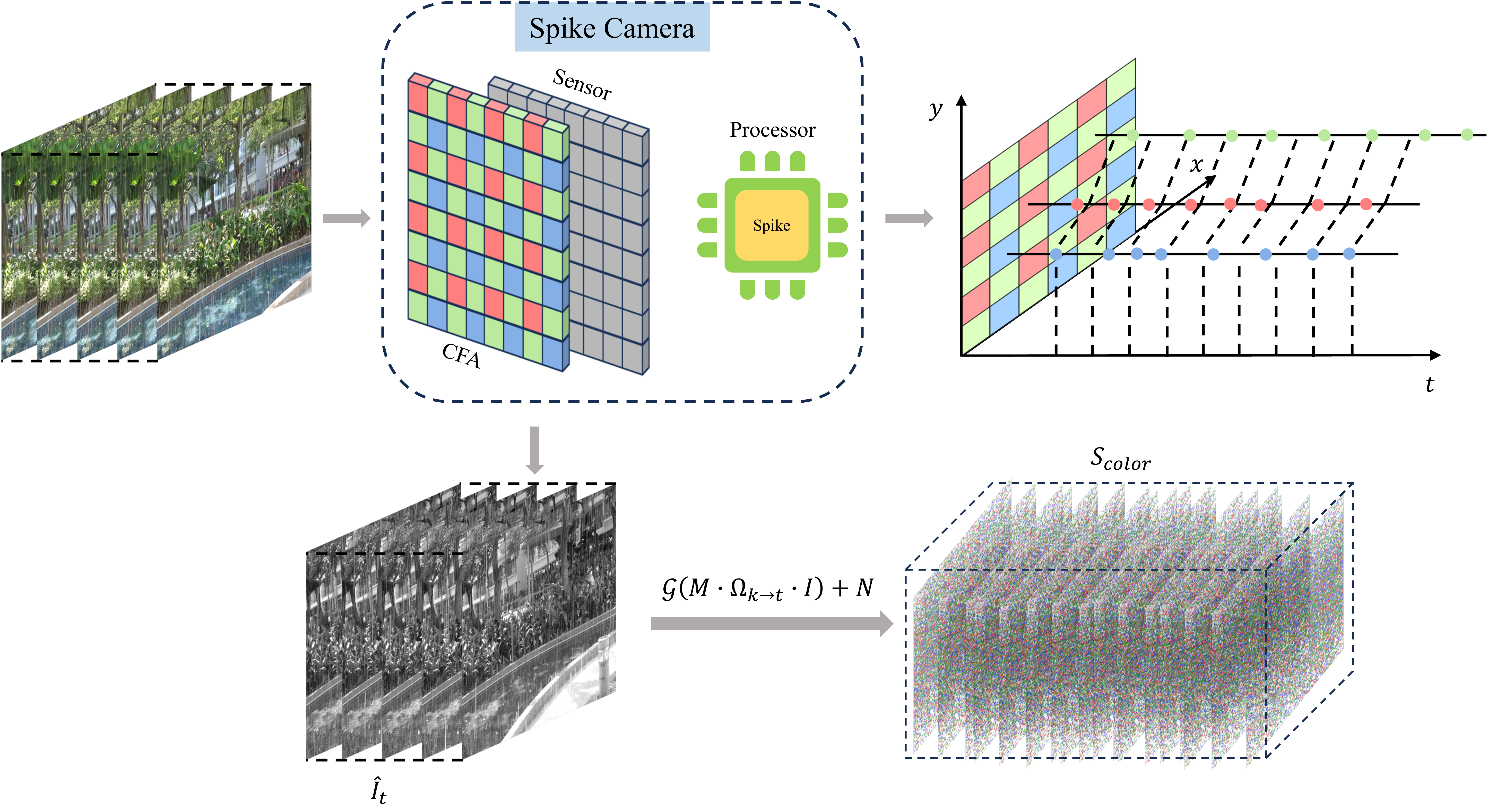}
	\vspace{-5pt}
	\caption{\textbf{Color Spike Stream with Bayer CFA.} A color spike stream is obtained using a Bayer pattern, which applies distinct binary masks to capture channel-specific spike signals, thereby enabling robust color reconstruction in dynamic environments.}
	\label{fig:cfa}
\end{figure}

\subsection{Color Spike Reconstruction}

\paragraph{Color Spike Stream with Bayer CFA.} To capture color information in dynamic scenes, a Bayer-pattern color filter array (CFA) is integrated with the spike camera, enabling the camera to generate distinct spike signals for each color channel at specific pixel locations. It can be defined as:
\begin{equation}
	M = M_r + M_g + M_b,
\end{equation}
where $M_r$, $M_g$, and $M_b \in \{0,1\}^{H \times W}$ are binary masks for the respective color channels, $H$ and $W$ are the image height and width. In high-speed scenes with short exposures, light intensity frames $\{\hat{I}_1, \hat{I}_2, \dots, \hat{I}_n\}$ captured by spike cameras are generated from dynamic scenes using a temporal window. The $t$-th latent light intensity frame $\hat{I}_t$ is modeled as:
\begin{equation}
	\hat{I}_t = M \cdot \Omega_{k \to t} \cdot I, \quad t \in \{1, 2, \dots, n\},
\end{equation}
where $I$ is the target light intensity to be reconstructed from the spike frames, and $\Omega_{k \to t}$ denotes the motion transformation matrix from the reference frame $k$ to the $t$-th frame. As shown in \cref{fig:cfa}, these noisy latent frames are processed by the spike camera mechanism to produce a color spike stream. Thus, Eq.~\eqref{con:spike_raw} can be modified to:
\begin{equation}
	S_{color} = \mathcal{G} \left( M \cdot \Omega_{k \to t} \cdot I \right) + N.
\end{equation}
The spike signals generated by each channel are recorded at different positions in $S_{\mathrm{color}}$.

\vspace{-10pt}
\paragraph{Spike Reconstruction Module.} The reconstruction module aims to extract color information and motion patterns from the color spike streams captured by the spike camera. This process involves handling noise in the spike data and predicting missing pixels for each color channel. The module addresses both the temporal and spatial challenges posed by the Bayer pattern and the binary nature of the spike stream, which includes the \textbf{Color Spike Representation (CSR)}, \textbf{Spike Noise Estimator (SNE)}, and \textbf{Color Correlation Prior (CCP)}.

Initially, the color spike streams are passed through the CSR module, which aligns the temporal features of each color channel, accounting for motion offsets between frames. By multiplying the spike frames with the Bayer mask for each color channel, we obtain frames with missing pixels. A multi-scale 3D convolution encoder processes these frames, extracting temporal features at different scales. These features are fused using an offset-sharing deformable convolution (OSDCN) module, which learns a shared offset for each color channel, aligning features and correcting for motion-induced misalignment across frames. The final output of CSR is a set of aligned feature blocks for each color channel, mathematically formulated as:
\begin{equation}
	\mathrm{Output} = \mathrm{Concat} \left( \mathrm{OSDCN} \left( \mathrm{3DEncoder} \left( M_c \cdot S \right) \right) \right),
\end{equation}
where $M_c \cdot S$ represents the separated spike stream by color channel, and $\mathrm{3DEncoder}(\cdot)$ is the multi-scale 3D convolution encoder.


A spike noise estimator (SNE) is employed to estimate noise in spike data, which is affected by mechanical and quantization noise. These noise sources degrade feature extraction quality, making accurate estimation crucial for high-quality detail recovery. The SNE module processes spike frames using a residual block-based encoder to capture noise features for each color channel. A decoder then generates a global noise map, characterizing noise across all channels. Then color correlation prior (CCP) module refines the reconstruction process by capitalizing on spatial and temporal correlation between color channels. Specifically, it concatenates the global noise features $N_{\mathrm{global}}$ from SNE with the feature blocks $F_r, F_g, F_b$ obtained from the CSR. 
This integration facilitates unified processing of color and noise features, enforcing spatiotemporal consistency, and leveraging interchannel dependencies to significantly enhance reconstruction fidelity.


 
\subsection{Spike-guided Sequences Deraining}

We propose a novel approach to video deraining by leveraging the SNN framework. This method extends the efficient spiking deraining network~\cite{ESDNet} to handle continuous video inputs, where temporal dependencies between frames are crucial for accurately removing rain streaks. By processing temporal sequences of frames, our method exploits correlations between consecutive frames, enhancing rain streak removal while preserving important video details.

\vspace{-10pt}
\paragraph{Spiking Residual Blocks.} 
Given a video input consisting of consecutive frames $I = \{ I_1, I_2, \dots, I_n \}$, where $n$ represents the total number of frames, we encode each frame $I_t$ into spike sequences using a biologically inspired encoding strategy. This converts each pixel's intensity into a temporal spike stream, representing changes in intensity over time. The Spiking Residual Blocks (SRB) are the core components of our deraining method, processing the temporal spike sequences to capture and remove rain streaks across frames. Each SRB consists of \textbf{Spike Convolution Unit (SCU)}, \textbf{Mixed Attention Unit (MAU)}, and \textbf{Residual Learning}.

The SCU converts the spiking representation into a temporal feature map by simulating membrane potential dynamics. It captures temporal variations and rain patterns by analyzing spike responses across consecutive frames. Leaky integrate-and-fire (LIF) neurons are utilized in the SCU to convert input signals into spike sequences, striking a balance between biological characteristics and computational complexity. The explicit dynamic equations for the LIF neurons can be formulated as:
\begin{equation}
	\left\{
	\begin{aligned}
	H^{t,n} & = U^{t-1,n} + \frac{1}{\tau} \left( X^{t,n} - \left( U^{t-1,n} - V_{\mathrm{reset}} \right) \right), \\
	S^{t,n} & = \Theta\left( H^{t,n} - V_{\mathrm{thr}} \right), \\
	U^{t,n} & = \left( \beta H^{t,n} \right) \odot \left( 1 - S^{t,n} \right) + V_{\mathrm{reset}} S^{t,n},
	\end{aligned}
	\right.
\end{equation}
where $H^{t,n}$ is the membrane potential of the neuron at time $t$ and state $n$, $S^{t,n}$ represents the firing state (1 if the neuron fires a spike, 0 otherwise), and $U^{t,n}$ is the membrane potential update. The neuronal membrane potential $H^{t,n}$ is determined by the membrane potential of the previous time step $U^{t-1,n}$, the input current $X^{t,n}$, and the time constant $\tau$. If the membrane potential exceeds the threshold $V_{\mathrm{thr}}$, the neuron emits a spike, setting $S^{t,n}$ to 1, and the membrane potential resets to $V_{\mathrm{reset}}$. Otherwise, the membrane potential continues to update and decay.

The mixed attention unit (MAU) is designed to improve the efficiency of spike processing by focusing on the most relevant parts of the input spike train. In traditional spike-based models, irrelevant or noisy spike events can distort learned features, reducing the model’s ability to focus on important temporal or spatial components of the signal. The MAU addresses this by applying spatial and temporal attention mechanisms, dynamically adjusting the model’s focus. Spatial attention operates on the spatial dimensions of the input (\textit{e.g.}, spatial locations within an image), while temporal attention selectively focuses on relevant time steps based on the temporal dynamics of the spike events.

Residual learning helps prevent issues such as information loss or gradient vanishing, which can occur in deep networks. By introducing skip connections, it ensures that low-level features are preserved and efficiently propagated across layers. Overall, Assuming that the output of the $n$-th SRB at the $t$-th time step is $I^{t,n}$, the calculated process of SRB can be depicted as:
\begin{equation}
	\left\{
	\begin{aligned}
	\hat{I}^{t,n}_1 & = \mathrm{SCU} \left( \mathrm{SCU} \left( I^{t,n-1} \right) \right), \\
	\hat{I}^{t,n}_2 & = \mathrm{tdBN} \left( \mathrm{Conv} \left( I^{t,n-1} \right) \right), \\
	I^{t,n} & = \mathrm{MAU} \left( \hat{I}^{t,n}_1 + \hat{I}^{t,n}_2 \right) + I^{t,n-1},
	\end{aligned}
	\right.
\end{equation}
where tdBN refers to threshold-dependent batch normalization (BN), a method designed for SNNs to address the limitations of traditional BN in handling sparse, time-dependent data in spiking neurons. 

To align with the characteristics of spike cameras, which capture continuous dynamic information at ultra-high temporal resolution, and to satisfy the spatiotemporal consistency requirements of our video deraining module, we construct a novel dynamic dataset, Rain100C. The rain streaks generated by our method adhere more closely to physical laws in terms of spatial distribution and brightness attenuation, thereby aligning well with the high-frequency data captured by spike cameras.

\section{Experimental Results}

We present comprehensive experiments conducted on commonly used benchmark datasets to evaluate the effectiveness of the proposed method. The experiments involve rigorous testing and analysis, providing a thorough evaluation of performance.

\subsection{Dataset Preparation}

\paragraph{\textsc{Rain100C} Dataset.} To obtain image sequences with continuous rain patterns more conveniently, we propose a rain pattern synthesis method based on traditional additive models. By quantitatively adjusting the physical properties of rain patterns (\textit{e.g.}, direction, length, width), we introduce a new paradigm for continuous rain pattern synthesis. Unlike traditional synthetic continuous rain patterns, our proposed model is not limited to similar backgrounds, fixed sizes, or software dependencies. Using our proposed continuous rain pattern synthesis method, we generated \textsc{Rain100C}, based on \textsc{Rain100H}~\cite{52} background images. For each background image, we generated 14 rain layer with continuous rain streaks, with the physical characteristics of the rain patterns varying across different backgrounds. For more details, please refer to the supplementary materials.

\vspace{-10pt}
\paragraph{\textsc{RainSynLight25} and \textsc{RainSynHeavy25} Datasets.} These two datasets~\cite{ESTIL} consist of 190 rainy/sunny video pairs in the training set and 27 pairs in the testing set. The rainy-day sequences were synthesized using probability models. \textsc{RainSynHeavy25} contains denser rain streaks than \textsc{RainSynLight25}, and the corresponding spike signals for both are generated by spike simulators~\cite{53}.

\vspace{-10pt}
\paragraph{\textsc{NTURain} Dataset.} This dataset~\cite{51} is divided into two groups: one captured with a slow camera and the other with a fast camera. It contains 25 pairs of rainfall sequences, including clear training versions and 8 pairs of test versions. For each rain-free video, Adobe After Effects (AE) synthesizes 3 or 4 rain layers with varying settings, which are then added to the video. The color spike stream for this dataset is also generated by a spike simulator~\cite{53}.

\begin{table*}
	\centering
	\setlength{\tabcolsep}{9pt}
	\footnotesize
	\begin{tabular}{lccccccccc}
	\toprule[1.1pt]
	\multirow{2}[2]{*}{Method} & \multirow{2}[2]{*}{Venue}
	& \multicolumn{2}{c}{\textsc{NTURain}} & \multicolumn{2}{c}{\textsc{RainSynLight25}} & \multicolumn{2}{c}{\textsc{RainSynHeavy25}} & \multicolumn{2}{c}{\textsc{Rain100C}} \\
	\cmidrule(lr){3-4} \cmidrule(lr){5-6} \cmidrule(lr){7-8} \cmidrule(lr){9-10}
	& & PSNR $\uparrow$ & SSIM $\uparrow$ & PSNR $\uparrow$ & SSIM $\uparrow$ & PSNR $\uparrow$ & SSIM $\uparrow$ & PSNR $\uparrow$ & SSIM $\uparrow$ \\
	\midrule
	SLDNet~\cite{SLDNet} & CVPR'20 & 34.26 & 0.956 & 28.73 & 0.843 & 18.80 & 0.529 & 33.42 & 0.947\\
	D3RNet~\cite{9} & CVPR'20 & 32.53 & 0.947 & 28.21 & 0.879 & 24.65 & 0.772 & 34.22 & 0.959 \\
	JDNet~\cite{54} & ACM MM'20 & 32.95 & 0.953 & 28.87 & 0.886 & 24.62 & 0.758 & 34.61 & 0.966\\
	S2VD~\cite{S2VD} & CVPR'21 & 37.37 & 0.968 & 33.84 & 0.925 & 27.77 & 0.818 & \underline{38.15} & \underline{0.972}\\
	TransWeather~\cite{57} & CVPR'22 & 33.35 & 0.955 & 29.06 & 0.902 & 25.95 & 0.797 & 36.96 & 0.969 \\
	MPEVNet~\cite{43} & CVPR'23 & 33.49 & 0.958 & 30.08 & 0.904 & 26.01 & 0.792 & - & - \\
	ESTIL~\cite{ESTIL} & TPAMI'23 & \underline{37.48} & \underline{0.970} & \textbf{36.12} & \textbf{0.963} & \underline{28.48} & \underline{0.824} & - & - \\
	GridFormer~\cite{58} & IJCV'24 & 32.96 & 0.954 & 28.45 & 0.896 & 24.66 & 0.724 & 36.28 & 0.963\\
	EHNet~\cite{59} & IJCV'24 & 34.11 & 0.962 & 30.63 & 0.920 & 26.28 & 0.811 & - & - \\
	\rowcolor{gray!20}
	SpikeDerain (Ours) & & \textbf{38.14} & \textbf{0.972} & \underline{34.48} & \underline{0.954} & \textbf{30.79} & \textbf{0.899} & \textbf{39.24} & \textbf{0.985} \\
	\bottomrule[1.1pt]
	\end{tabular}
	\vspace{-5pt}
	\caption{\textbf{Quantitative comparisons on three public datasets and Rain100C.} Best and second-best results are highlighted in \textbf{bold} and \underline{underlined}.}
	\label{Quantitative}
\end{table*}

\subsection{Implementation Details and Metrics}

\paragraph{Implementation Details.}
During training, we use the Adam optimizer within the PyTorch framework. Initially, the reconstruction module is trained separately, with the CCP modules sharing the same parameters across iterations. The number of input spike frames $N$ is set to 39, and we randomly crop the spike frames into $64 \times 64$ patches, with a batch size of 8. Data augmentation is applied by randomly flipping the input frames horizontally and vertically. After completing the reconstruction module training, we freeze the module and train the overall framework using an Adam optimizer with a batch size of 12. The learning rate is set to $1 \times 10^{-3}$, and a cosine annealing strategy is used to gradually decrease the learning rate to $1 \times 10^{-7}$.

Due to the lack of corresponding spike streams on \textsc{NTURain}, \textsc{RainSynLight25}, and \textsc{RainSynHeavy25}, we used a spike simulator~\cite{53} combined with a Bayer Pattern CFA to generate corresponding color spike streams for training our model.

\vspace{-10pt}
\paragraph{Metrics.} We introduce peak signal-to-noise ratio (PSNR) and structural similarity index (SSIM) as quantitative evaluation metrics, computed in the $\mathrm{Y}$ channel of the $\mathrm{YCbCr}$ color space. To evaluate the efficiency of our method, we also calculate the energy consumption of the model, following the previous works~\cite{ternary}.

\subsection{Comparisons with State-of-the-Art Methods}

We have compared our network with several state-of-the-art methods. Among these, JDNet~\cite{54} is single-image deraining methods, MPEVNet~\cite{43} and EHNet~\cite{59} are event-driven methods, and TransWeather~\cite{57} and GridFormer~\cite{58} are general models designed to handle various weather conditions.

\vspace{-10pt}
\paragraph{Quantitative Results on Synthetic Data.} \cref{Quantitative} presents the quantitative results of our model compared to other advanced methods on three publicly available datasets and the proposed \textsc{Rain100C}. Our model outperformed all other methods on all four datasets in terms of both PSNR and SSIM. For example, on \textsc{NTURain}, our model achieved a PSNR of 38.14 dB and an SSIM of 0.972, showing a 0.66 dB improvement in PSNR over suboptimal results, ESTIL~\cite{ESTIL}. On \textsc{RainSynLight25}, our model achieved PSNR and SSIM values of 34.48dB and 0.954, second only to ESTIL~\cite{ESTIL}, and far exceeding the latest event-driven models, EHNet~\cite{59} and MPEVNet~\cite{43}. On \textsc{RainSynHeavy25}, which features a denser distribution of rain streaks, our model achieved PSNR and SSIM values of 30.79 dB and 0.899, respectively, representing an 8\% and 9\% improvement over suboptimal results, ESTIL~\cite{ESTIL}. For \textsc{Rain100C}, which contains a wider variety of rain pattern physical attributes and richer backgrounds, our method also achieved significantly better results compared to other state-of-the-art methods. These results emphasize the effectiveness and robustness of our method across various rainfall levels and datasets, setting new standards for video deraining tasks.

\begin{figure*}
	\centering
	\includegraphics[width = \linewidth]{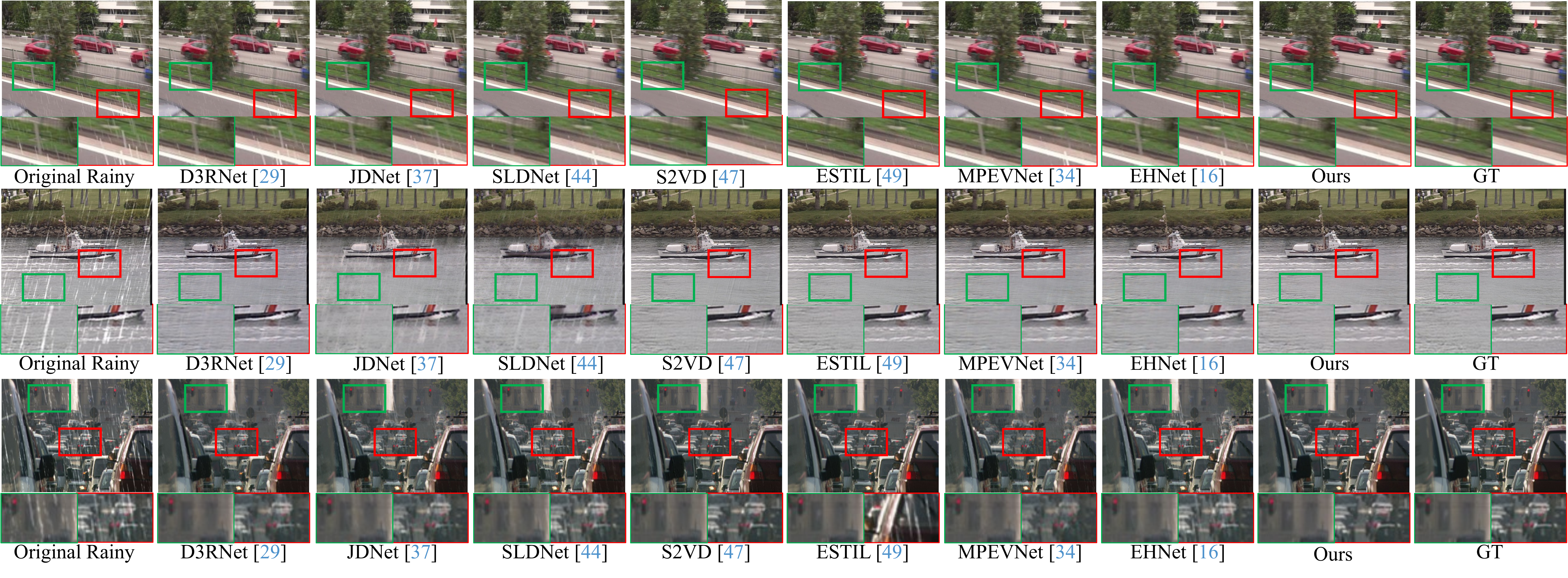}
	\vspace{-15pt}
	\caption{\textbf{Qualitative comparison on \textsc{NTURain} (Top), \textsc{SynHeavy25} (Middle) and \textsc{SynLight25} (Bottom).} The green box highlights thicker rain streaks, while the red box shows more challenging interlaced rain patterns. Zoom-in for better visualization.}
	\label{fig:qualitative}
\end{figure*}

\vspace{-10pt}
\paragraph{Qualitative Results on Synthetic Data.} \cref{fig:qualitative} presents a qualitative analysis of results on \textsc{NTURain} and \textsc{SynHeavy25}, which has more complex rain patterns. From the comparison, it is clear that D3RNet~\cite{9} and JDNet~\cite{54} remove only a small number of rain streaks in high-speed scenes, with many streaks remaining. This aligns with their poor performance in the quantitative results. MPEVNet~\cite{43} and EHNet~\cite{59} perform better in high-speed scenarios by introducing event streams but still face challenges in different scenarios. S2VD~\cite{S2VD} and ESTIL~\cite{ESTIL} are classic video deraining methods that effectively remove heavy rain streaks but still leave some subtle rain streaks. Among all comparison methods, our approach produces the best results, completing the restoration of detailed textures while extracting high-speed motion features.


\subsection{Ablation Studies}
To better understand the factors contributing to the superior performance of our method, we conducted ablation studies on \textsc{RainSynHeavy25}.

\begin{table}
	\centering
	\setlength{\tabcolsep}{6pt}
	\footnotesize
	\begin{tabular}{ccccc}
	\toprule[1.1pt]
	{Time steps} & {PSNR $\uparrow$} & {SSIM $\uparrow$} & {FLOPs (G) $\downarrow$} & {Energy (uJ) $\downarrow$} \\
	\midrule
	3 & 30.30 & 0.895 & \textbf{8.479} & \textbf{1.0565 $\times$ 10$^5$} \\
	4 & 30.58 & 0.896 & 8.578 & 1.0690 $\times$ 10$^5$ \\
	\rowcolor{gray!20}
	5 & \textbf{30.79} & \textbf{0.899} & 8.678 & 1.0814 $\times$ 10$^5$ \\
	6 & 30.47 & 0.894 & 8.777 & 1.0939 $\times$ 10$^5$ \\
	7 & 29.72 & 0.885 & 8.877 & 1.1064 $\times$ 10$^5$ \\
	\bottomrule[1.1pt]
	\end{tabular}
	\vspace{-5pt}
	\caption{\textbf{Impact of different time steps on \textsc{RainSynHeavy25}.} All time steps are configured during training.}
	\label{time_step}
\end{table}

\vspace{-10pt}
\paragraph{Effectiveness of Different Time Steps.} \cref{time_step} shows the performance at various time steps, revealing that longer time windows generally improve image restoration quality by enabling the network to capture richer spatiotemporal features. However, as the time step grows, the computational overhead and energy consumption also increase. Specifically, performance gains begin to the plateau beyond five time steps, indicating diminishing returns relative to energy cost. To balance performance and efficiency, we selected a time step of 5 as the default.

\begin{figure}
	\centering
	\includegraphics[width = 0.98\linewidth]{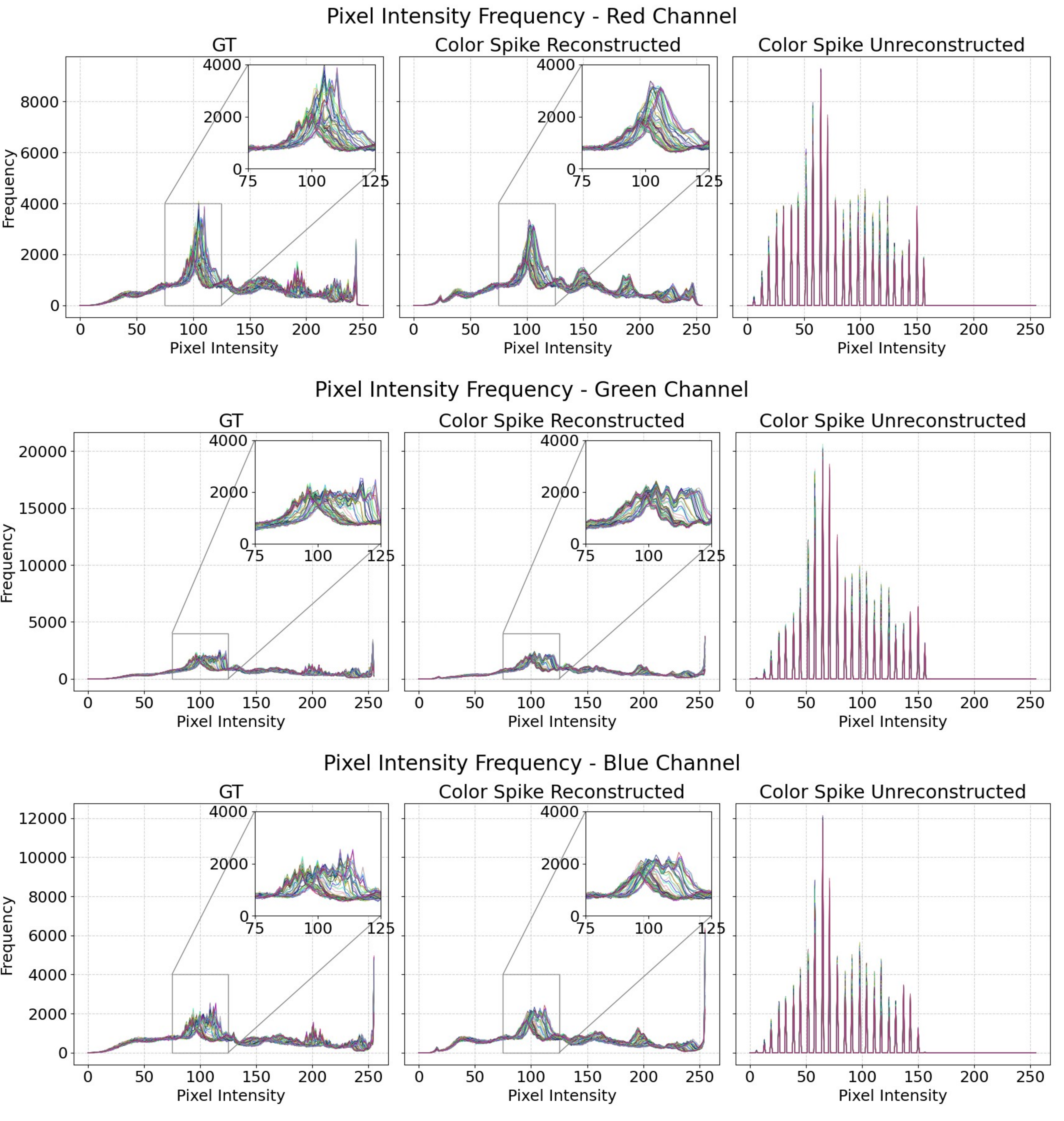}
	\vspace{-5pt}
	\caption{\textbf{Pixel Intensity Frequency Analysis.} From top to bottom and from left to right, represent the different pixel frequency analyses of the original image, reconstructed image, and original spike stream in the three channels, respectively.}
	\label{fig:pixel_analysis}
\end{figure}

\vspace{-10pt}
\paragraph{Effectiveness of Color Spike Reconstruction.} As shown in \cref{fig:pixel_analysis}, the reconstructed pixel intensity histograms closely align with the ground truth in all three channels, indicating that the proposed Color Spike Reconstruction method preserves both the shape of the overall distribution and fine details. This high degree of overlap demonstrates minimal distortion, maintaining spatial continuity and inter-channel consistency without introducing noticeable color shifts or artifacts. Consequently, the reconstructed images effectively retain the original color characteristics, confirming the robustness and accuracy of the method.

\begin{table}
	\centering
	\setlength{\tabcolsep}{10pt}
	\footnotesize
	\begin{tabular}{lccc}
	\toprule[1.1pt]
	{Reconstruction Method} & {PSNR $\uparrow$} & {SSIM $\uparrow$} & {LPIPS $\downarrow$} \\
	\midrule
	TFI~\cite{18} & 28.32 & 0.843 & 0.171 \\ 
	TFP~\cite{18} & 30.34 & 0.889 & 0.112 \\
	BSN~\cite{31} & 24.87 & 0.714 & 0.477 \\
	\rowcolor{gray!20}
	SpikeDerain (Ours) & \textbf{30.79} & \textbf{0.899} & \textbf{0.104} \\
	\bottomrule[1.1pt]
	\end{tabular}
	\vspace{-5pt}
	\caption{\textbf{Impact of different reconstruction methods on \textsc{RainSynHeavy25}.}}
	\label{reconstruct}
\end{table}

\vspace{-10pt}
\paragraph{Effectiveness of Different Reconstruction Methods.} To evaluate the performance of our reconstruction module, we compared it with several common spike reconstruction methods. \cref{reconstruct} shows the PSNR/SSIM/learned perceptual image patch similarity (LPIPS) values of the final rain removal results using different reconstruction methods. To match the color spike stream of the Bayer pattern, we reconstructed the image into three channels using the CFA mask. The comparison reveals that the traditional TFP method preserves color information less effectively, while the BSN method disrupts color correlation between channels. Our reconstruction method fully preserves color information and avoids color distortion during subsequent processing.

\begin{table}
	\centering
	\setlength{\tabcolsep}{3pt}
	\footnotesize
	\begin{tabular}{lccccccc}
	\toprule[1.1pt]
	{Model} & {BN} & {tdBN} & {CU} & {SCU} & {PSNR $\uparrow$} & {SSIM $\uparrow$} & {LPIPS $\downarrow$} \\
	\midrule
	ANN & \CIRCLE & \Circle & \CIRCLE & \Circle & 30.21 & 0.889 & 0.126 \\
	SNN & \CIRCLE & \Circle & \Circle & \CIRCLE & 30.71 & 0.896 & 0.112 \\
	ANN \textit{w/} tdBN & \Circle & \CIRCLE & \CIRCLE & \Circle & 29.96 & 0.876 & 0.132 \\
	\rowcolor{gray!20}
	SNN \textit{w/} tdBN & \Circle & \CIRCLE & \Circle & \CIRCLE & \textbf{30.79} & \textbf{0.899} & \textbf{0.104} \\
	\bottomrule[1.1pt]
	\end{tabular}
	\vspace{-5pt}
	\caption{\textbf{Quantitative ablation study on different components of our network using \textsc{RainSynHeavy25}.} CU replaces the spiking neuron in SCU with an ANN neuron.}
	\label{design}
\end{table}

\vspace{-10pt}
\paragraph{Effectiveness of SNN.} We further examined the impact of SNNs on deraining performance via a series of ablation studies. As shown in \cref{design}, incorporating spiking neurons and tdBN consistently enhances the overall metrics. Notably, the SCU excels in capturing temporal dynamics and addressing rain streaks. Compared to conventional convolutions, SNN-based operations paired with tdBN yield higher PSNR/SSIM and lower LPIPS, indicating improved visual fidelity. These results underscore the effectiveness of our spiking architecture in exploiting the asynchronous nature of spike data for robust rain removal.

\section{Conclusion}
In this work, we propose SpikeDerain, a single-modal video deraining framework that leverages color spike streams captured by a novel spike camera. Unlike traditional cameras limited by fixed exposure and motion blur, and event cameras that capture only sparse brightness changes, our spike camera continuously records absolute brightness and color information at ultra-high temporal resolution. This enables our network to effectively reconstruct detailed color images and accurately separate rain streaks from the background, thereby delivering superior deraining performance.

\section*{Acknowledge}
This work was supported in part by the National Natural Science Foundation of China under Grant 62271361 and 62301213, and the Hubei Provincial Key Research and Development Program under Grant 2024BAB039.

{
 \small
 \bibliographystyle{ieeenat_fullname}
 \normalem
 \bibliography{SpikeDerain}
}


\clearpage
\setcounter{page}{1}
\setcounter{section}{0}
\setcounter{figure}{0}
\setcounter{equation}{0}
\maketitlesupplementary

\section{Detail of Rain100C}
\label{sec:rationale}
\subsection{Motivation of Rain100C}
The motivation for building this dataset stems from some challenges in image processing tasks in current rainfall scenarios. In existing public datasets, the synthesis of raindrops or rain streaks often lacks sufficient realism, especially in depicting the continuity of raindrops and natural motion patterns. Due to simplification or excessive discretization in the generation process of rain streaks in many existing datasets, the synthesized images exhibit obvious local fractures, disjointed tails, or raindrop motion patterns that do not conform to physical laws. These inherent defects often become factors that affect the performance of algorithms in practical applications.

\begin{figure}[h]
	\centering
	\includegraphics[width = \linewidth]{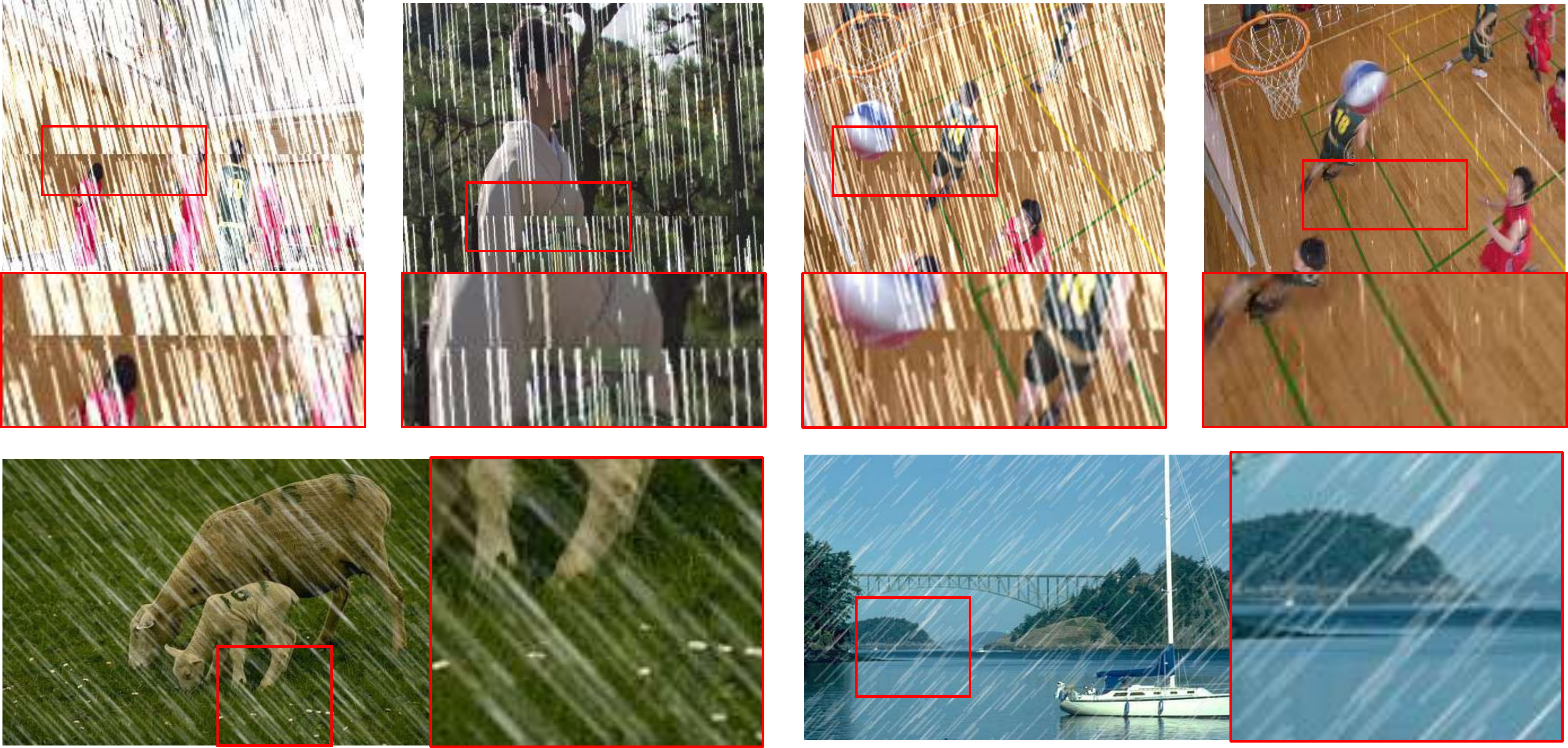}
	\caption{\textbf{Visual comparison between existing datasets and our method for generating rain streaks}. There is a phenomenon of discontinuous rain streaks in the existing dataset, while the rain streaks generated by this method are visually more natural and continuous.}
\label{fig:Rain100C}
\end{figure}

In \cref{fig:Rain100C}, the top row shows the rain patterns synthesized from the existing public dataset, and the bottom row shows the rain patterns generated by this method. It can be observed that in existing datasets, there are often phenomena such as discontinuous rain patterns and abrupt tails (as shown in the red rectangular box), which make it difficult to reflect the coherent motion of real raindrops visually; The rain patterns generated by this method significantly improve these discontinuities, with smoother raindrop tails and more consistent physical laws in spatial distribution and brightness attenuation. This feature helps to provide more realistic and diverse training samples for image rain removal, object detection, and other computer vision tasks in rainy scenes, thereby improving the robustness and generalization ability of the algorithm in practical application scenarios.

\subsection{Method of Construction}
We propose a method that synthesizes continuous rain streaks by leveraging the underlying physics of rainfall. Our approach is summarized in the following theoretical formulation:

Firstly, an initial noise map $ N(x,y) $ is generated via uniform random sampling. This noise is then processed using a threshold function to obtain:
\begin{equation}
	N'\left(x,y\right) = \begin{cases}
	N\left(x,y\right), & \mathrm{if } N\left(x,y\right) \geq 255 - v, \\
	0, & \mathrm{otherwise},
	\end{cases}
\end{equation}
where $ v $ is a parameter controlling the noise level. This step ensures that only high-intensity noise is preserved, thereby better reflecting the local highlights of raindrops in the imaging process.

Subsequently, we construct a motion blur kernel $ K(x,y) $ to simulate the trailing effect observed when raindrops are in high-speed motion. This kernel is designed based on the physical parameters of raindrops, such as the length $ L $, width $ w $, and tilt angle $ \theta $. The kernel construction can be conceptualized as initially forming a vertical line at the center of the kernel and then applying a rotation:
\begin{equation}
	K'\left(x,y\right) = \mathrm{rotate}\left(L\left(x,y\right), \theta\right),
\end{equation}
with normalization such that:
\begin{equation}
	\sum_{x,y} K'\left(x,y\right) = 1,
\end{equation}
ensuring that the overall brightness of the image remains unchanged after convolution.

By convolving the processed noise map with the motion blur kernel, we obtain the rain streak map $ R(x,y) $:
\begin{equation}
	R\left(x,y\right) = \left(K * N'\right)\left(x,y\right).
\end{equation}

A linear fusion model is employed to synthesize the rainy image. The original image $I(x,y)$ is overlaid with the rain streak map $R(x,y)$ using a weighting coefficient $ \beta $:
\begin{equation}
	I_{\mathrm{rain}} \left(x,y\right) = I \left(x,y\right) + \beta R \left(x,y\right),
\end{equation}
where $\beta$ controls the opacity of the rain streaks, ensuring that the original image details are retained while a natural rainy effect is achieved.

Furthermore, to capture the continuous motion of raindrops, spatial displacement parameters $ (h, w) $ are introduced. For consecutive frames $t$, the rain streak map is smoothly translated, which is expressed as:
\begin{equation}
	I_{\mathrm{rain}}^{(t)}\left(x,y\right) = I\left(x,y\right) + \beta R\left(x - h_t, y - w_t\right).
\end{equation}
This continuous spatial translation ensures temporal coherence in the generated data, more accurately reflecting the dynamic characteristics of raindrops in natural rainfall scenarios.

\section{Theoretical Energy Consumption of SNNs}
In the ablation studies of this paper, we compared the performance of models with different time steps, including the energy consumption calculation of spike neural networks. In this work, we calculated the theoretical energy consumption of different models. 

The energy advantage of spiking neural networks (SNNs) primarily stems from their event-driven computation. In SNNs, synaptic operations occur only when a neuron’s membrane potential exceeds a threshold and a spike is triggered. This greatly reduces the actual number of operations compared to traditional neural networks. The theoretical energy consumption of an SNN can be estimated by counting the total number of synaptic operations (SOPs) that occur over all time steps. This can be expressed as:
\begin{equation}
	N_{\mathrm{SOP}} = s \times T \times A,
\end{equation}
where $s$ is the average sparsity of neuron activations (\textit{i.e.}, the proportion of neurons that actually fire), $T$ is the total number of simulation time steps, an $A$ is the number of additive operations in the corresponding artificial neural network (ANN).

Furthermore, the total energy consumption of the SNN is given by the sum of the energy consumed by these synaptic operations and the energy used for spike generation:
\begin{equation}
	E_{\mathrm{SNN}} = N_{\mathrm{SOP}} \times E_{\mathrm{SOP}} + N_{\mathrm{sign}} \times E_{\mathrm{sign}},
\end{equation}
where, $E_{\mathrm{SOP}}$ denotes the energy consumption per synaptic addition operation, and $E_{\mathrm{sign}}$ represents the energy consumed by a spike (Sign operation) triggered by an LIF neuron. $N_{\mathrm{sign}}$ is the total number of spikes triggered. In contrast, a typical floating-point operation (FLOP) in a conventional ANN requires approximately 12.5pJ of energy. By leveraging low-energy additions and the sparse firing mechanism, SNNs exhibit a significant advantage in overall energy consumption.

This model, which is based on the statistics of the actual number of operations and the energy cost of different operation types, facilitates the theoretical quantification of the low-energy characteristics of SNNs and provides strong theoretical support for their implementation in efficient, resource-constrained devices.

\section{More Qualitative Results}
\begin{figure}[h]
	\centering
	\includegraphics[width = \linewidth]{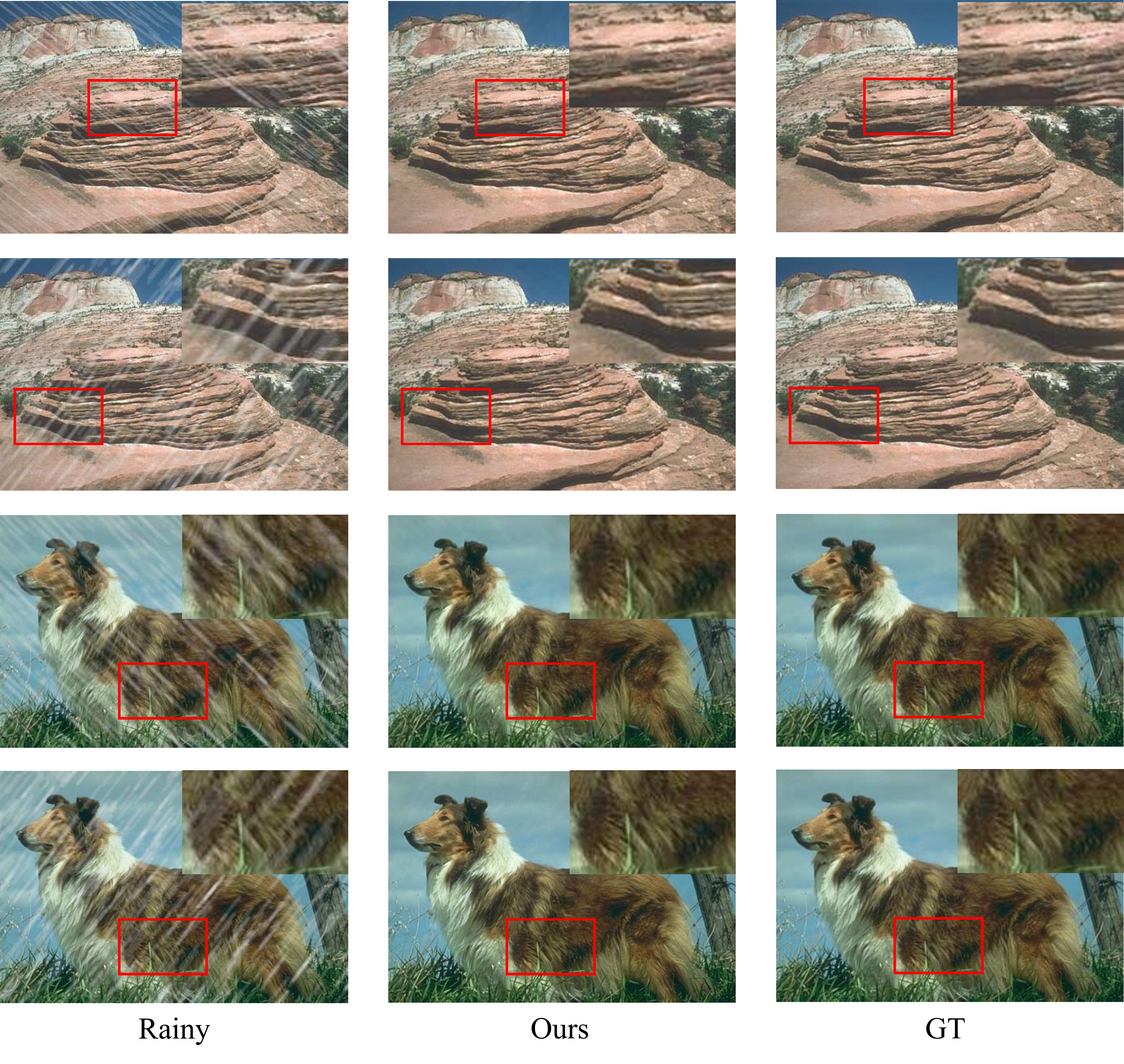}
	\caption{\textbf{Qualitative results on Rain100C.} Zoom-in for better visualization.}
	\label{fig:qualitative_rain100c}
\end{figure}

\begin{figure*}
	\centering
	\includegraphics[width = \linewidth]{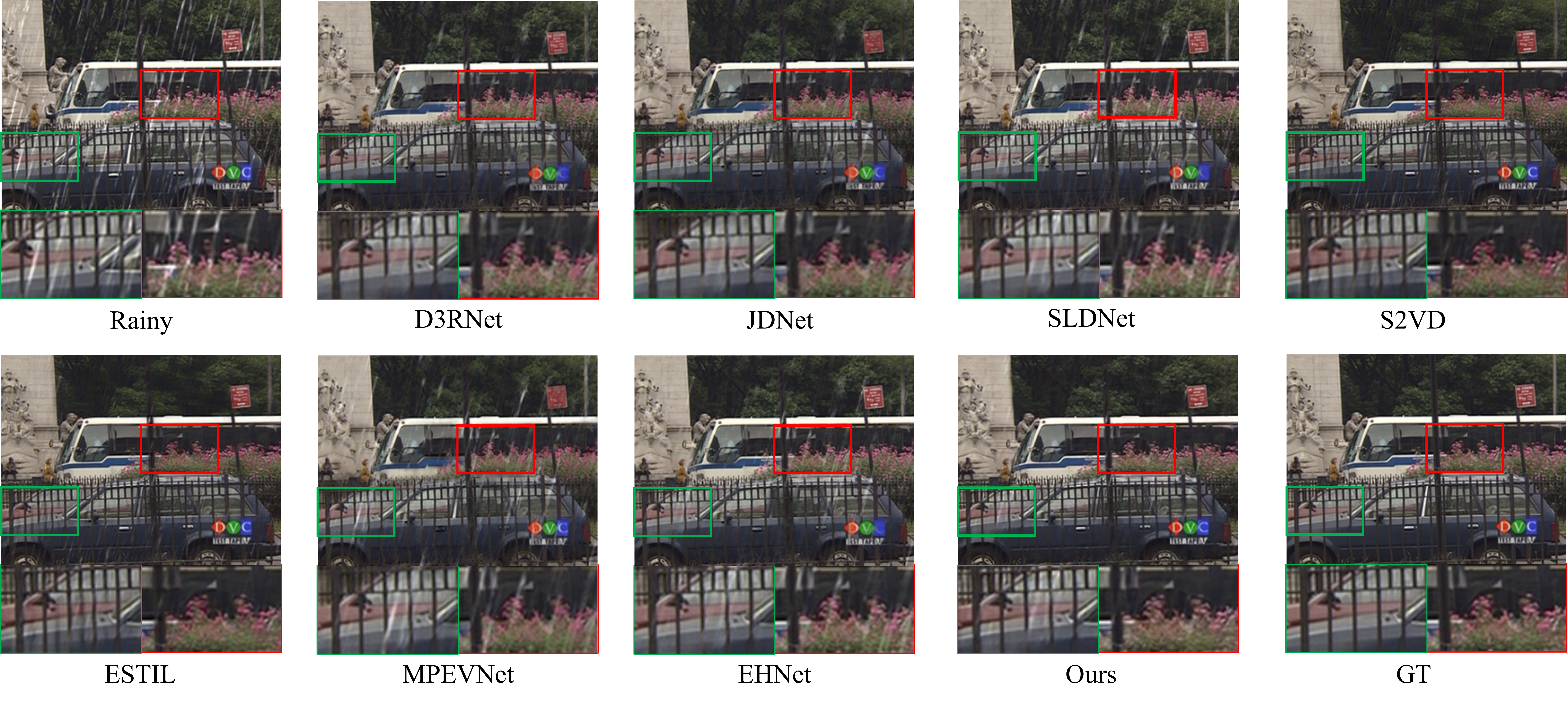}
	\caption{\textbf{Qualitative comparison on SynLight25.} Zoom-in for better visualization.}
	\label{fig:qualitative_light}
\end{figure*}

\begin{figure*}
	\centering
	\includegraphics[width = \linewidth]{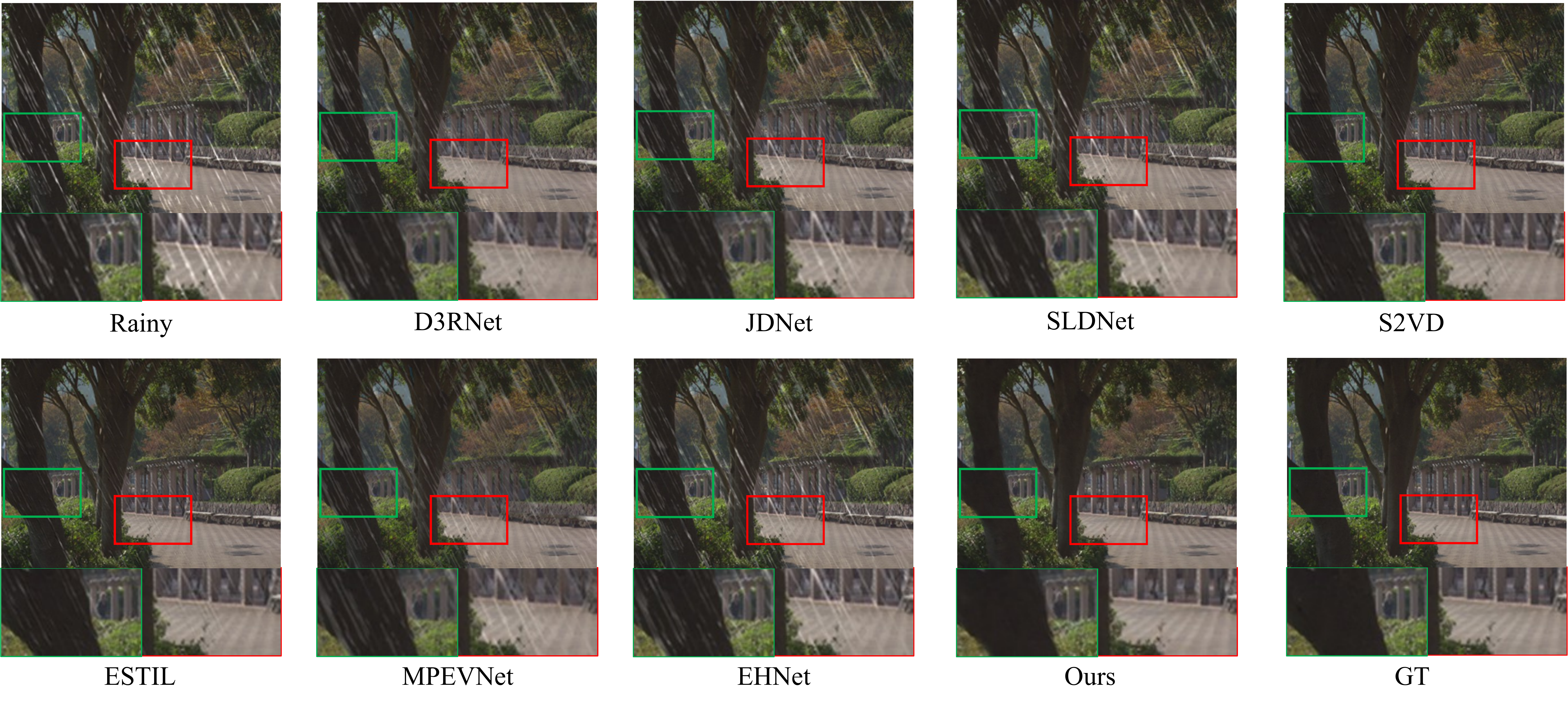}
	\caption{\textbf{Qualitative comparison on SynLight25.} Zoom-in for better visualization.}
	\label{fig:qualitative_light2}
\end{figure*}

\begin{figure*}
	\centering
	\includegraphics[width = \linewidth]{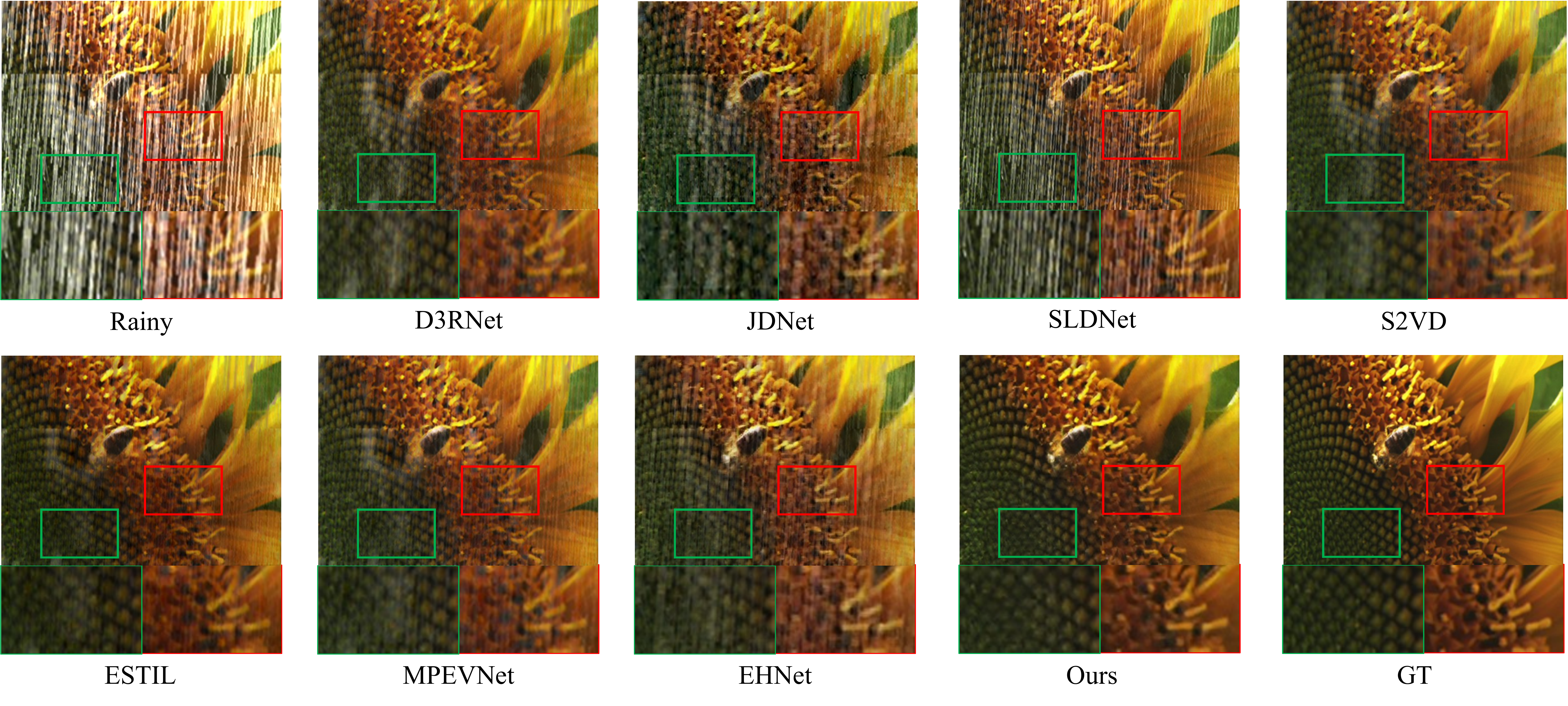}
	\caption{\textbf{Qualitative comparison on SynHeavy25.} Zoom-in for better visualization.}
	\label{fig:qualitative_heavy1}
\end{figure*}

\begin{figure*}
	\centering
	\includegraphics[width = \linewidth]{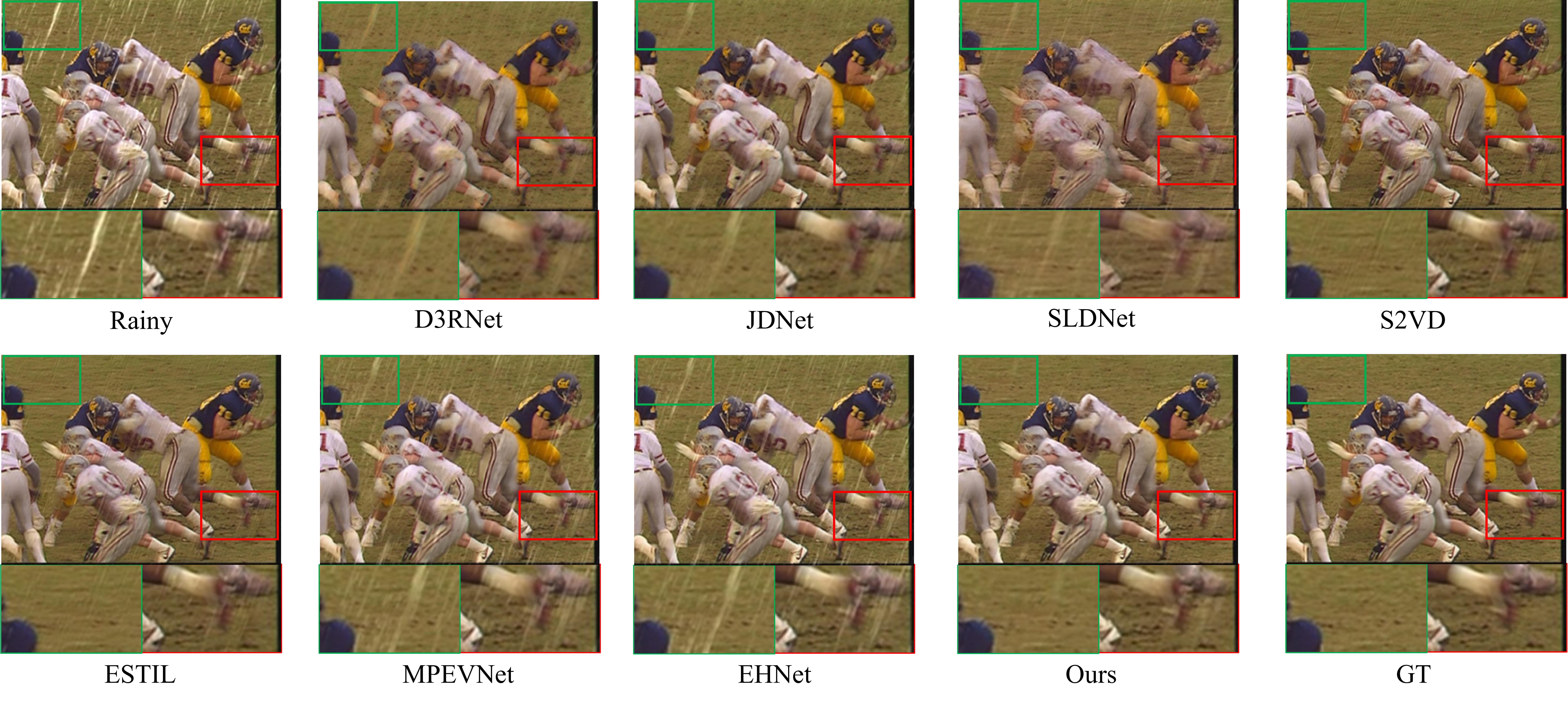}
	\caption{\textbf{Qualitative comparison on SynHeavy25.} Zoom-in for better visualization.}
	\label{fig:qualitative_heavy2}
\end{figure*}

We present more qualitative comparisons on various datasets, shown in \cref{fig:qualitative_rain100c} to \cref{fig:qualitative_heavy2}. It can be observed that our proposed method outperforms other state-of-the-art methods, with the best effects of rain removal and detail restoration.

\end{document}